\documentclass[mlmain]{jmlr}

\usepackage{longtable}
\usepackage{booktabs}
\usepackage[load-configurations=version-1]{siunitx} 

\theorembodyfont{\upshape}
\theoremheaderfont{\scshape}
\theorempostheader{:}
\theoremsep{\newline}

\jmlrvolume{}
\firstpageno{1}
\editors{List of editors' names}

\jmlryear{2024}
\jmlrworkshop{Symmetry and Geometry in Neural Representations}

\title{Visualizing Loss Functions as Topological Landscape Profiles}

 \author{
 \Name{Caleb Geniesse\nametag{\thanks{Equal contribution.}}} \Email{cgeniesse@lbl.gov}\\
 \addr Lawrence Berkeley National Laboratory
 \AND
 \Name{Jiaqing Chen$^*$} \Email{jchen501@asu.edu}\\
 \addr Arizona State University 
 \AND
 \Name{Tiankai Xie$^*$} \Email{txie21@asu.edu}\\
 \addr Arizona State University
 \AND
 \Name{Ge Shi} \Email{geshi@ucdavis.edu}\\
 \addr University of California, Davis
 \AND
 \Name{Yaoqing Yang} \Email{yaoqing.yang@dartmouth.edu}\\
 \addr Dartmouth College
 \AND
 \Name{Dmitriy Morozov} \Email{dmorozov@lbl.gov}\\
 \addr Lawrence Berkeley National Laboratory
 \AND
 \Name{Talita Perciano} \Email{tperciano@lbl.gov}\\
 \addr Lawrence Berkeley National Laboratory
 \AND
 \Name{Michael W.~Mahoney} \Email{mmahoney@stat.berkeley.edu}\\
 \addr ICSI, LBNL, and UC Berkeley 
 \AND
 \Name{Ross Maciejewski} \Email{rmacieje@asu.edu}\\
 \addr Arizona State University 
 \AND
 \Name{Gunther H.~Weber} \Email{ghweber@lbl.gov}\\
 \addr Lawrence Berkeley National Laboratory
}

\begin{document}
\maketitle

\begin{abstract}
    In machine learning, a loss function measures the difference between model predictions and ground-truth (or target) values. For neural network models, visualizing how this loss changes as model parameters are varied can provide insights into the local structure of the so-called loss landscape (e.g., smoothness) as well as global properties of the underlying model (e.g., generalization performance). While various methods for visualizing the loss landscape have been proposed, many approaches limit sampling to just one or two directions, ignoring potentially relevant information in this extremely high-dimensional space. This paper introduces a new representation based on topological data analysis that enables the visualization of higher-dimensional loss landscapes. After describing this new topological landscape profile representation, we show how the shape of loss landscapes can reveal new details about model performance and learning dynamics, highlighting several use cases, including image segmentation (e.g., UNet) and scientific machine learning (e.g., physics-informed neural networks). Through these examples, we provide new insights into how loss landscapes vary across distinct hyperparameter spaces: we find that the topology of the loss landscape is simpler for better-performing models; and we observe greater variation in the shape of loss landscapes near transitions from low to high model performance. 
\end{abstract}


\section{Introduction}
\label{sec:introduction}

A central aim of machine learning \citep{simonyan2014very,he2016deep,krizhevsky2017imagenet,vaswani2017attention,devlin2018bert,liu2019roberta} is to learn the underlying structure of data. This learning process is governed by a \emph{loss function}, denoted as $\mathcal{L}(\theta)$, where $\theta$ is the set of parameters (or weights) defining, e.g., a neural network. The loss function measures the difference between the outputs of a neural network and ground-truth values. In this way, the loss reflects how good (or bad) the current weights are at making correct predictions and how to adjust these weights during training. Given the important role that the loss function plays during learning, examining it with respect to a neural network's weights---by visualizing the so-called \emph{loss landscape}---can provide valuable insights into both network architecture and learning dynamics \citep{goodfellow2014qualitatively,im2016empirical,li2018visualizing,yao2020pyhessian,martin2021implicit,martin2021predicting,yang2022evaluating,yang2021taxonomizing,zhou2023three,sakarvadia2024mitigating,khan2024sok}. Indeed, the loss landscape has been essential for understanding other aspects of deep learning, including generalizability \citep{cha2021swad,yang2021taxonomizing} and robustness \citep{kurakin2016adversarial,djolonga2021robustness,yang2022generalized}.
In addition, the loss landscape has been characterized in the context of scientific machine learning, e.g., to understand why different physics-informed architectures and loss functions are often brittle, exhibiting failure modes, and are hard to optimize \citep{krishnapriyan2021characterizing,rathore2024challenges,xie2024evaluating}. 

Despite its promise and appeal, \emph{loss landscape visualization} is a complex and often bespoke process. Indeed, exploring and extracting insights from a loss landscape---which is inherently high-dimensional, with as many dimensions as the number of parameters in the model---is challenging to do, especially when trying to visualize directly on a two-dimensional screen. Most efforts to date have focused on projecting the loss function down to one or two dimensions. \citet{goodfellow2014qualitatively} proposed a random-direction-based approach, where model parameters are interpolated along a one-dimensional path to see how the loss changes. \citet{im2016empirical} later introduced an extension of this method which involves projecting the loss landscape onto a two-dimensional space using barycentric interpolation between triplets of points and bilinear interpolation between quartets of points. \citet{li2018visualizing} continued improving the resolution of loss landscapes by introducing filter-wise normalization to remove the scaling effects incurred by previous approaches. A more sophisticated approach to visualizing the loss landscape leverages the Hessian to define more relevant directions along which the model can be interpolated. More recently, \citet{yao2020pyhessian} used the top two Hessian eigenvectors as directions, thereby capturing more important changes in the underlying loss landscape. While various methods have been proposed, most applications have limited sampling to just one or two directions. Importantly, by restricting the sampling of loss landscapes to two dimensions, whether it be using random or Hessian-based directions, we ignore potentially informative information captured by additional dimensions (e.g., the eigenvectors associated with the dominant eigenvalues of the Hessian matrix). 

Towards characterizing higher-dimensional loss landscapes, here we take inspiration from topological data analysis (TDA). Specifically, we use a merge tree to encode the critical points of an $n$-dimensional neural network loss landscape, and we represent the merge tree as a topological landscape profile. The merge tree allows us to capture important features in an arbitrary-dimensional loss landscape; and by using the topological landscape profile, we are able to re-represent this information in two dimensions. Note, we first explored applications of TDA in our previous work \citep{xie2024evaluating}. There, we focused on quantifying loss landscapes and developing new topology-based metrics, but the sampling was limited to two dimensions. Here, we focus on developing a new visual representation of the loss landscape which allows us to visualize higher-dimensional loss landscapes. We demonstrate the utility of our new topological landscape profile representation by exploring higher-dimensional loss landscapes, i.e., sampling along more directions and representing these higher-dimensional subspaces as topological landscape profiles. This approach allows us to extract more information from the additional dimensions we consider. While our approach technically can work with arbitrary dimensional loss landscapes, in practice we are limited by sampling. As such, here we limit ourselves to three and four-dimensional loss landscapes. 

We demonstrate the versatility of our new topological profile representations of loss landscapes and our complementary visualization tool through several use case scenarios. Through these examples, we show the many different ways our tool can be used to extract insights about neural network models based on our topological landscape profiles and by comparing loss landscapes across different hyperparameters. In doing so, we also provide new insights into how loss landscapes vary across distinct hyperparameter spaces, finding that (1) the topology of loss landscapes is simpler for better-performing models, and (2) this topology often exhibits greater variability near transitions from low to high model performance. For example, for the scientific machine learning models we study here, we observe a sharp transition from low to high error as one of the physical parameters is increased. We find that models with lower error have smoother and more funnel-like loss landscapes, whereas models with higher error have flatter (but rougher) and more bowl-like loss landscapes. Along this transition from low to high error, models have more variably shaped loss landscapes (i.e., different shapes are observed across different random seeds).

\section{Background}
\label{sec:background}

\subsection{Topological Data Analysis}
\label{subsec:tda}

Topological data analysis (TDA) aims to reveal the global underlying structure of data. TDA is particularly useful for studying high-dimensional data or functions, where direct visualization (in two or three dimensions) is inherently not possible. We leverage ideas and algorithms from TDA to study the global structure of the loss function—that is, the shape of the so-called loss landscape. Much of TDA is based on the more general idea of ``connectedness.'' In the context of a loss function, we are interested in the number of minima (i.e., unique sets of parameters for which the loss is locally minimized) and how ``prominent'' they are (i.e., measuring how many other sets of neighboring parameters have a higher loss than the parameter set that minimizes the loss function). Such information can be obtained from a persistence diagram (i.e., captured by the zero-dimensional persistent homology) and the so-called merge tree.

A \emph{merge tree} \citep{contour_tree, heine2016survey} tracks connected components of sub-level sets $L^-(v) = \{ x \in D; x \le v\}$ as a threshold, $v$, is increased. The merge tree encodes changes in the loss landscape as nodes in a tree-like structure. The local minima are represented by degree-one nodes, which are connected to other local minima through a single saddle point. The saddle points connecting different minima are represented by degree-three nodes (each connecting two local minima and one other saddle point). Loss functions often display many shallow local minima with low barriers (i.e., the value difference between the minima and the connecting saddle point is small) corresponding to ``short-lived'' connected components that merge quickly with other connected components. 

In our work, we use the merge tree to extract the underlying structure of a loss landscape. We then use this extracted information to construct our topological landscape profile representations. Since the merge tree can be computed for an arbitrary dimensional loss landscape, we can use it to construct our representation for higher-dimensional loss landscapes, which would otherwise be difficult to visualize.

\subsection{Topological Landscape Profiles}
\label{subsec:topological_profiles}

To enable the visualization of higher-dimensional loss landscapes, we introduce a new topological landscape profile representation that captures the minima and saddle points encoded by merge trees. This work builds upon \citet{oesterling2013visualizing}, who first introduced the idea of representing high-dimensional data clusters (and their nesting) as hills in a landscape, where the height, width, and shape of each hill encodes the coherence, size, and stability of each cluster. To construct the landscape profile, they first use a merge tree to encode the distribution (or density) of the data points. They then use this merge tree to construct the landscape profile, by representing maxima in the merge tree as hills in the landscape, where the size and shape of each hill are determined by characteristics like persistence and the number of points along the corresponding branch. In the context of loss functions, we are more interested in minima than maxima, so here we introduce a new version of this topological landscape profile, using the metaphor of valleys (or basins) rather than hills.

\section{Methods}
\label{sec:methods}

\begin{figure*}[th]
    \centering	
    \includegraphics[width=\linewidth]{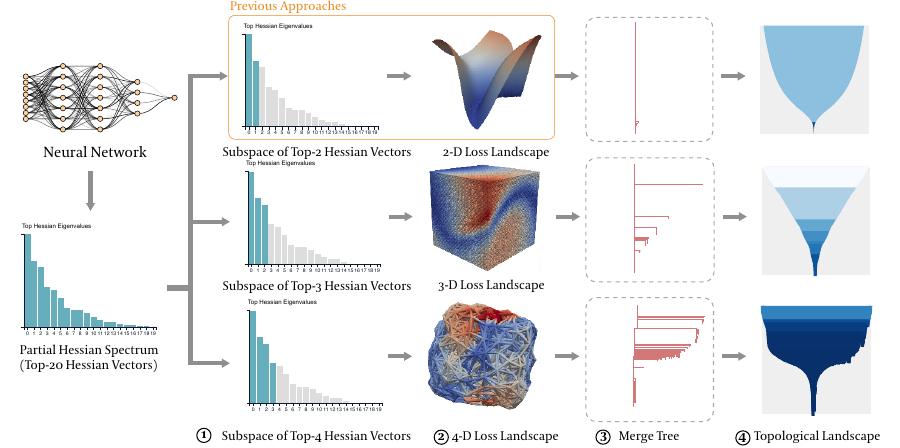}
    \caption{
    Our topological landscape profiles enable the visualization of higher-dimensional loss landscapes by capturing their underlying shape (or topology). Here we show loss landscapes based on the top $n$ Hessian eigenvectors. See Section~\ref{sec:methods} for details. 
    }
    \label{fig:analytics_framework}
    \vspace{-4mm}
\end{figure*}

To construct our new topological landscape profile representations, we build on traditional loss landscape sampling approaches and leverage tools from TDA to capture the underlying shape (or topology) of the sampled loss landscapes. First, we select $n$ vectors ($n \leq m$) to define an $n$-dimensional subspace (Figure~\ref{fig:analytics_framework}.1), where $m$ is the number of parameters in the model. We then sample a set of points from this subspace, where each point corresponds to a distinct set of parameters. We evaluate the loss for each set of parameters and represent the set of points (and their associated loss values) as an unstructured grid (Figure~\ref{fig:analytics_framework}.2). We then compute a merge tree to capture the topology of the $n$-dimensional loss landscape (Figure~\ref{fig:analytics_framework}.3), and we construct our final topological landscape profile based on this merge tree (Figure~\ref{fig:analytics_framework}.4). In this section, we go into more detail about each of these steps.

\subsection{Loss Landscape Construction and Representation}
\label{sec:methods_nd_loss_landscape_representations}

In this work, we limit our analysis to Hessian-based loss landscapes. We calculate the top $n$ Hessian eigenvectors using PyHessian \citep{yao2020pyhessian} (Figure~\ref{fig:analytics_framework}.1) and then sample along the subspace spanned by these directions (Figure~\ref{fig:analytics_framework}.2). The idea is that by using the eigenvectors associated with the top $n$ largest eigenvalues, we can visualize the most significant local loss fluctuations for a given model. Given the $n$ orthogonal directions, we generalize the approach taken by \citet{li2018visualizing} by expanding the subspace beyond two dimensions. Formally, we perturb trained model parameters along the $n$ directions and evaluate the loss $\mathcal{L}$ as follows:
\begin{equation}
  \label{eq:projection_general}
  f(\alpha_1 ... \alpha_n) = \mathcal{L} (\theta + \Sigma_{i=1}^{n}\alpha_i\delta_i)  ,
\end{equation}
where $\alpha_1...\alpha_n$ are the coordinates in the $n$-dimensional subspace, $\delta_i$ is the $i$-th direction in that subspace, and $\theta$ is the original model. As such, each coordinate corresponds to a point associated with a computed loss value, and the collection of loss values forms an $n$-dimensional loss landscape. In this work, we use an equally spaced grid by taking each $\alpha_i$ to be the set of equally spaced integers between $0$ and $r$, where $r$ is the resolution of each dimension in the grid. Here we use $r = 41$, such that the center of the grid corresponds to the original model, i.e., $\Sigma_{i=1}^{n}\alpha_i\delta_i = 0$.

Given an $n$-dimensional loss landscape, we can represent the sampled points as an unstructured grid, where each vertex in the grid is associated with $n$ coordinates and a scalar loss value. Before we can characterize how the loss changes throughout the landscape (i.e., as parameters are perturbed from one vertex to the next), we need to define the spatial proximity (or connectivity) of vertices in the grid based on the similarity of their coordinates. Here we use a scalable, approximate nearest neighbor algorithm to construct a neighborhood graph representation of the loss landscape \citep{dong2011efficient}. The \emph{neighborhood graph}, proposed by \citet{jaromczyk1992relative}, of a dataset $D$ is a graph $G = (D,E)$ where two points $u$ and $v$ are connected by an edge $(u, v) \in E$ if they are $similar$. Here we focus on the $k$-nearest neighbor graph, where each point is connected to the $k$ most similar points. We also use a $symmetric$ version of this graph, where points are only considered neighbors if each point is a neighbor of the other. In this case, an edge $(u, v)$ is pruned from the graph if $u$ is not one of the $k$ nearest points to $v$, or vice versa. We note that this approach involves selecting an appropriate value for the $k$ parameter. Here we use $k = 4 \times n$, such that the connectivity is similar to the spatial proximity of pixels in an image (i.e., each pixel having $k = 8$ neighbors, corresponding to the left, right, top, bottom, and all four corners).

\subsection{Topological Structures and Landscape Profiles}
\label{sec:methods_topological_structures_and_landscape_profiles}

After defining the subspace and computing the loss landscape, we perform topological data analysis to extract and summarize the most important features. In this work, we use a merge tree to extract key information from the loss landscape, which we then use to define our topological landscape profile. We compute the merge tree for each loss landscape using the Topology ToolKit (TTK), developed by \citet{ttk2021overview}. 

Given a merge tree, we then construct the topological landscape profile using the method proposed by \citet{oesterling2013visualizing}. In this representation, each branch (in the merge tree) ending in a local minimum is represented by a basin (in the landscape profile), and each sub-branch ending in a saddle point is represented as a sub-basin, below which other basins are placed. In either case, each basin (or sub-basin) is represented by a set of rectangles encoding the cumulative size of the branch (or sub-branch), from bottom to top, such that the top of the basin is as wide as the number of points found along the corresponding branch in the merge tree.

\begin{figure}
    \centering	
    \includegraphics[width=1.00\columnwidth]{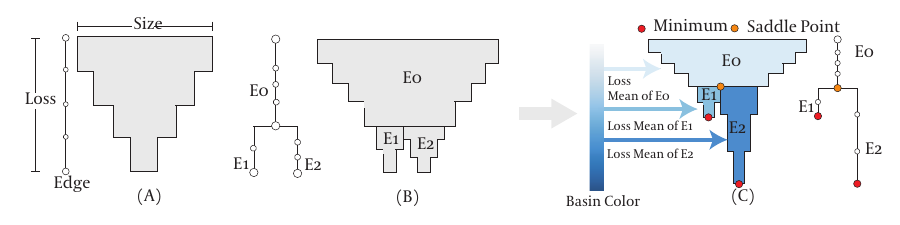}
    \caption{
    Representing the merge tree as a topological landscape profile. In (A) we show a single basin corresponding to a merge tree with a single branch, and in (B) we show multiple basins corresponding to multiple branches. In (C) we color the basins based on their average loss.
    }
\label{fig:topological_landscape_profile_construction}
    \vspace{-4mm}
\end{figure}

We introduce this topological landscape profile representation of loss functions to effectively capture more information from higher-dimensional loss landscapes, in such a way that can still be visualized. While this topological representation and the merge tree used to create it both capture important features of the high-dimensional space, it also discards some important information by design. Here, we reincorporate some of this discarded information back into our representation, for example, by using the loss values to color the different basins. As shown in Figure~\ref{fig:topological_landscape_profile_construction}.C, we compute the average loss across the points in each basin, and we use darker blues to represent lower average loss values. Thus, deeper basins are represented by a darker blue color, evoking the idea of deeper ocean depths. In addition to coloring the topological landscape profile, we also annotate the basins with the critical points, including saddle points (orange dots) and minima (red dots). Interestingly, the distribution (or density) of saddle points and minima reflects local characteristics of the loss landscape, such as locally sharp or locally flat.

\section{Empirical Evaluation}
\label{sec:experiments_results}

\subsection{Visualizing Different Physical Constraints}
\label{sec:experiments_results_pinn}

In our first evaluation, we look at a set of physics-informed neural network (PINN) models trained to solve simple convection problems \citep{krishnapriyan2021characterizing}. Here we aim to investigate the PINN’s soft regularization and how it helps (or fails to help) the optimizer find an optimal solution to a seemingly simple convection problem. We show how the shape and complexity of our topological landscape profiles change as a physical ``wave speed'' parameter is increased and the PINN fails to solve this seemingly simple physical problem. Specifically, we consider the one-dimensional convection problem, a hyperbolic partial differential equation that is commonly used to model transport phenomena:
\begin{align}\label{eq:PINN_1}
    \frac{\partial u}{\partial t} + \beta_{\text{}} \frac{\partial u}{\partial x} = 0,\ x \in \Omega,\ t \in [0, T] 
\end{align}
 \vspace{-6mm}
\begin{align}\label{eq:PINN_2}
    u(x, 0) = h(x),\ x \in \Omega 
\end{align}
where $\beta_{\text{}}$ is the convection coefficient and $h(x)$ is the initial condition. The general loss function for this problem is
\begin{align}\label{eq:PINN_3}
    L(\theta) = \frac{1}{N_{u}}\sum_{i=1}^{N_{u}} (\hat{u}-u_0^i)^2 + \frac{1}{N_{f}}\sum_{i=1}^{N_{f}} \lambda_i (\frac{\partial \hat{u}}{\partial t} + \beta_{\text{}} \frac{\partial \hat{u}}{\partial x})^2 + L_B
\end{align}
where $\hat{u} = N N (\theta, x, t)$ is the output of the NN, and $L_B$ is the boundary loss. While increasing the physical wave speed parameter, $\beta_{\text{}}$, should not necessarily make this a harder problem to solve, it can make PINN models harder to train. Interestingly, \citet{krishnapriyan2021characterizing} related these failure modes to changes in the corresponding loss landscape, showing that it becomes increasingly complicated, such that optimizing the model becomes increasingly difficult. Here we explore these failure modes in more detail using three-dimensional and four-dimensional Hessian-based loss landscapes, finding more variability in the shape of loss landscapes near the transition between high and low-performing models.

\begin{figure*}[t!]
  \centering
  \includegraphics[width=\linewidth]{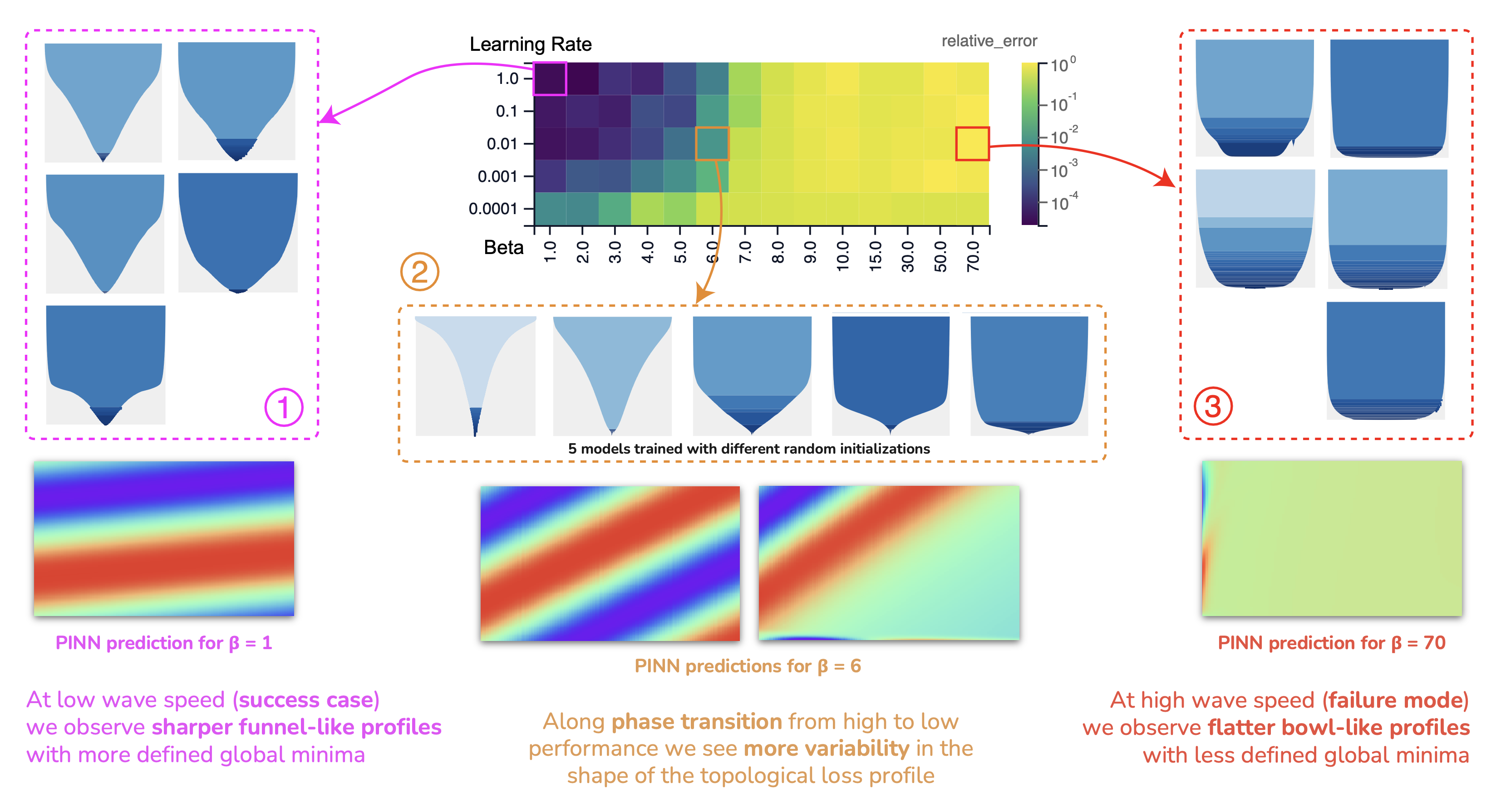}
  \vspace{-10mm}
  \caption{
   Analyzing the loss function of a physics-informed neural network (PINN) trained to solve simple physical convection problems. See Section~\ref{sec:experiments_results_pinn} for details.
  }
  \label{fig:teaser}
    \vspace{-4mm}
\end{figure*}

In Figure~\ref{fig:teaser}, we show a heat map corresponding to the average relative error across different values of the physical wave speed parameter and across different learning rates. Interestingly, we observe that the error increases with this physical parameter, but more slowly for higher learning rates. The smallest learning rate displays higher error rates even for smaller values of the physical parameter. When looking at the loss landscapes, we observe consistently more funnel-like loss landscapes for the smaller values of $\beta$, corresponding to lower error (Figure~\ref{fig:teaser}.1). In contrast, we observe a consistently more bowl-like loss landscape for the larger values of $\beta$, corresponding to higher error (Figure~\ref{fig:teaser}.3). The funnel-like landscapes correspond to when the PINN models find a physically reasonable solution, albeit constrained to a smaller space of solutions by the physical wave speed parameter. In other words, since the solution is constrained by the physical parameter, perturbing the model results in a faster increase in the loss, given that the physical problem is no longer satisfied. In contrast, the more bowl-like landscapes correspond to the failure to find a reasonable solution, such that perturbing the model does not immediately change the already high loss. Note, the landscapes corresponding to these failure modes also include more saddle points and are otherwise more complex.

To verify that these representations are stable across different model initializations, we show five different landscapes for each hyperparameter configuration, corresponding to the same model trained using different random seeds. We see the landscapes look similar across different random seeds for the low and high values of the physical wave speed parameter. Moreover, we observe more variation in the loss landscapes near the transition from low to high error (Figure~\ref{fig:teaser}.2). This suggests that while the error starts to increase near the transition, only some of the models are failing whereas other may be finding physically reasonable solutions, as indicated by the funnel-like loss landscapes. 

In Figure~\ref{apd:first}.\ref{fig:case-pinn-high-dim-1} we compare the topological landscape profiles based on three- and four-dimensional loss landscapes. An important insight here is that, in higher dimensions, we observe many more critical points and that the basins in the much spikier landscape can be mapped back to the wider basins in the topological landscape profiles based on the three-dimensional loss landscapes. Overall the global shape of the topological landscape profile looks similar when comparing the same random seed. Moreover, this highlights an important feature of our new representations---the ability to visualize higher-dimensional loss landscapes, i.e., sampling along more than just one or two dimensions.

\subsection{Visualizing Loss Landscapes Over Training}
\label{sec:experiments_results_unet}

In our second evaluation, we explore how loss landscapes change throughout training and across different learning rates. To do this, we study UNet models with a learnable CRF-RNN layer \citep{avaylon2022adaptable} trained on the Oxford-IIIT Pet dataset \citep{parkhi2012cats}. We trained the models using five different random seeds across seven different learning rates for 30 epochs. For each checkpoint, we computed two-dimensional loss landscapes based on the top two Hessian eigenvectors. The model was perturbed using a distance of 0.01 and layerwise normalization was adopted \citep{li2018visualizing}.

\begin{figure*}[b!]
    \centering	
    \includegraphics[width=\linewidth]{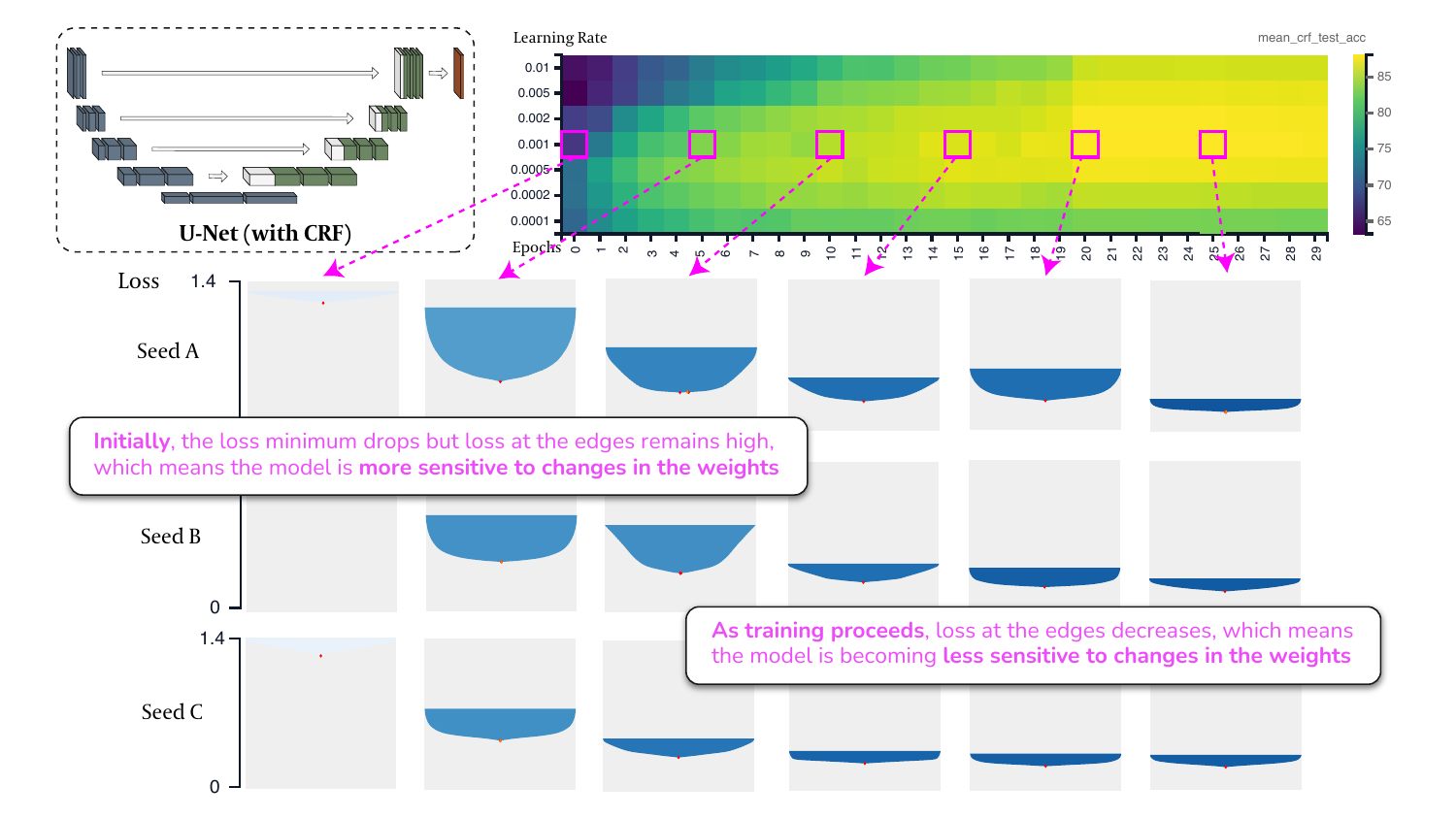}
    \vspace{-6mm}
    \caption{
    Loss landscapes over training for UNet models with a CRF layer trained on the Oxford-IIIT Pet dataset. See Section~\ref{sec:experiments_results_unet} for details.
    }
    \label{fig:case-unet}
    \vspace{-4mm}
\end{figure*}

In Figure~\ref{fig:case-unet} and Figure~\ref{apd:second}.\ref{fig:case-unet-vertical}, we show the same heat map corresponding to average test accuracy over training and across different learning rates. We observe that the test accuracy improves over training, with some variation across the different learning rates. In Figure~\ref{fig:case-unet}, we consider how the loss landscape changes over training. When looking at the loss landscapes for three different random seeds, after zooming in, we observe an initially shallow loss landscape but with the global minimum at a much higher loss compared to the end of training. As training proceeds, we see that the global minimum becomes lower, but the basin itself becomes deeper with the edges remaining at much higher loss values. As the global minimum continues to drop, we also observe additional flattening of the basin, such that all points have a much lower loss compared to the beginning of training. Interestingly, the flat basin at a much higher loss corresponds to a phase of learning where perturbing the model in any one direction doesn't really increase the already high loss. After five epochs, the much deeper basin reflects a less stable model, where perturbing the model results in relatively higher loss. As training proceeds, we observe a flattening of the basin, which means the model becomes more stable, as perturbations result in smaller changes in loss. In Figure \ref{apd:second}.\ref{fig:case-unet-vertical}, we consider how the loss landscape changes across different learning rates. When looking at the loss landscapes for three different random seeds, after zooming in, we observe consistent variation in the depth and shape of the loss landscape as the learning rate is varied. This variation is also reflected in the test accuracy scores shown in the heat map. Interestingly, we observe deeper basins when the learning rate is too small or too big, indicating that the trained models are less stable compared to those with shallower basins.

\section{Conclusion and Future Work}
\label{sec:conclusion_future_work}

In this paper, we introduced a new topological landscape profile representation of neural network loss landscapes. To demonstrate the many different ways this new representation of loss landscapes can be used, we explored several different machine learning examples, including image segmentation (e.g., UNet-CRF) and scientific machine learning (e.g., PINNs). Along the way, we provided new insights into how loss landscapes vary across distinct hyperparameter spaces, finding that the topology of the loss landscape is simpler for better-performing models and that this topology is more variable near transitions from low to high model performance. Moreover, by using a merge tree to extract the most important features from a computed loss landscape, we are able to construct a new representation encoding these features. By separating this new representation from the original space in which the loss landscape was sampled, our approach opens up the door to visualizing higher-dimensional loss landscapes.

While we only explore up to four dimensions here, our approach can be extended to any number of dimensions. The limiting factor is sampling, which requires exponentially many more resources as the number of dimensions increases. However, future advances towards more efficient sampling could be combined with our current approach to reveal the higher-dimensional structure of loss functions. Complementary advances in sampling more global loss landscapes (combining multiple independently trained models) could also benefit from our new representations. In that case, we would expect to see more distinct basins in our topological landscape profiles.

\section{Acknowledgments}

This work was supported by the U.S. Department of Energy, Office of Science, Advanced Scientific Computing Research (ASCR) program under Contract Number DE-AC02-05CH11231 to Lawrence Berkeley National Laboratory and Award Number DE-SC0023328 to Arizona State University (“Visualizing High-dimensional Functions in Scientific Machine Learning”). This research used resources at the National Energy Research Scientific Computing Center (NERSC), a U.S. Department of Energy, Office of Science, User Facility under NERSC Award Number ASCR-ERCAP0026937.

\bibliography{main}

\newpage 

\appendix
\label{sec:appendix}

\section{Visualizing Different Physical Constraints}\label{apd:first}

\begin{figure*}[h]
    \centering	
    \includegraphics[width=\linewidth]{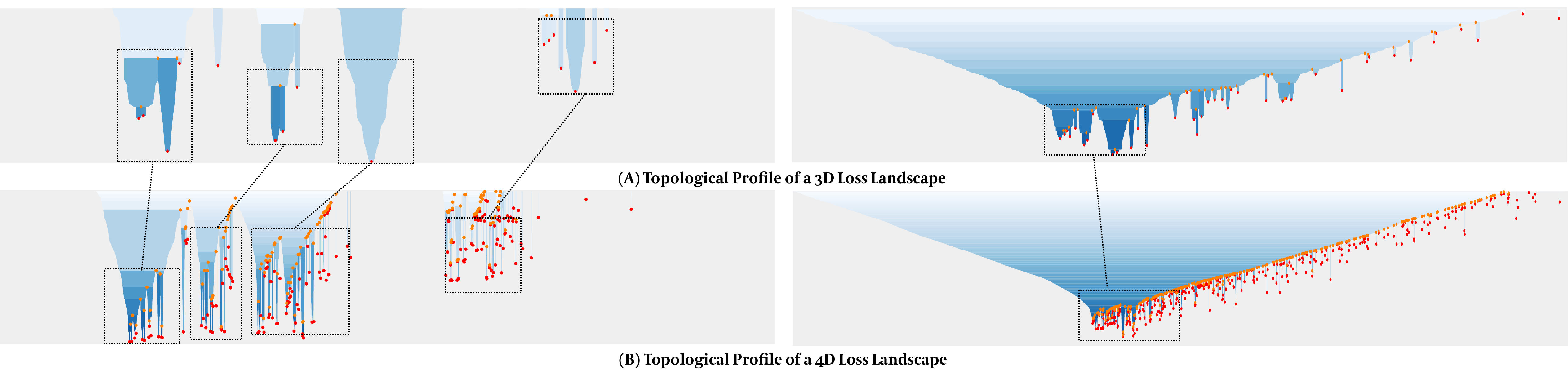}
    \caption{
    Comparing topological landscape profiles based on (A) three-dimensional and (B)~four-dimensional loss landscapes. See Section~\ref{sec:experiments_results_pinn} for details.
    }
    \label{fig:case-pinn-high-dim-1}
\end{figure*}

In Figure \ref{apd:first}.\ref{fig:case-pinn-high-dim-1}, we show topological landscape profiles for two different random initializations (from left to right) of a physics-informed neural network (PINN). Note, that the landscapes look different for the different random initializations because we are looking at a model corresponding to the transition from low to high error. Overall the global shape of the topological landscape profile looks similar when comparing the same random seed across three- and four-dimensional loss landscapes. In four dimensions, we observe many more critical points and that the basins in the much spikier landscape can be mapped back to the wider basins in the topological landscape profiles based on the three-dimensional loss landscapes. These visualizations also highlight an important feature of our topological landscape profile representations---the ability to visualize higher-dimensional loss landscapes.

\newpage

\section{Visualizing Loss Landscapes Over Training}\label{apd:second}

\begin{figure}[h]
    \centering	
    \includegraphics[width=\linewidth]{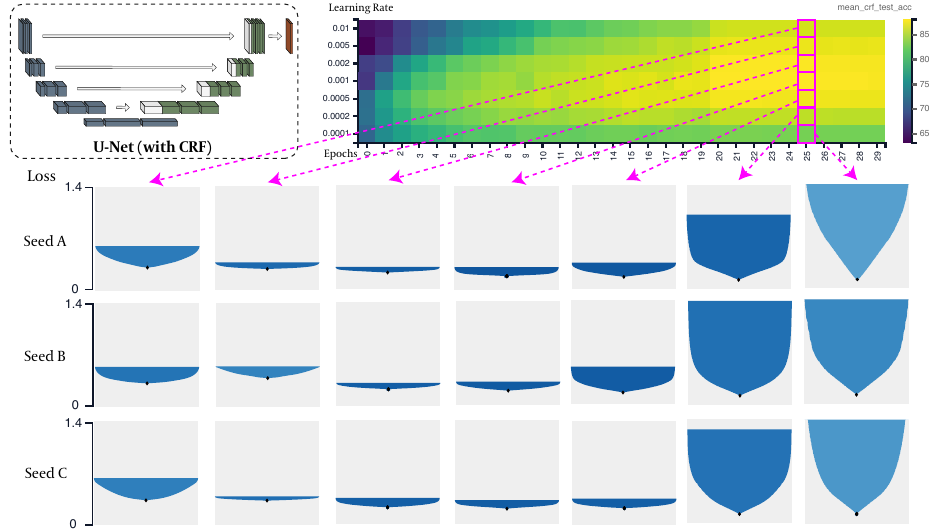}
    \caption{
    Loss landscapes across learning rates for UNet models with a CRF layer trained on the Oxford-IIIT Pet dataset. See Section~\ref{sec:experiments_results_unet} for details.
    }
    \label{fig:case-unet-vertical}
\end{figure}

In Figure \ref{apd:second}.\ref{fig:case-unet-vertical}, we show how the loss landscape changes across different learning rates. When looking at the loss landscapes for three different random seeds, after zooming in, we observe consistent variation in the depth and shape of the loss landscape as the learning rate is varied. See Section~\ref{sec:experiments_results_unet} for details.

\end{document}